\documentclass[11pt]{article}

\usepackage{acl}

\usepackage{times}
\usepackage{latexsym}

\usepackage{amsmath}
\usepackage{amssymb}
\usepackage{booktabs}
\usepackage{multirow}
\usepackage{tabularx}
\usepackage[most]{tcolorbox}
\usepackage{subcaption}

\newcolumntype{C}{>{\centering\arraybackslash}X}
\definecolor{oursgreen}{RGB}{0,100,0}
\newtcolorbox{casebox}[1][]{%
  enhanced,
  breakable,
  colback=gray!3,
  colframe=gray!55,
  boxrule=0.35pt,
  arc=1.5mm,
  left=1mm,
  right=1mm,
  top=1mm,
  bottom=1mm,
  #1
}
\newcommand{\wrongthink}[1]{\textcolor{red!75!black}{#1}}
\newcommand{\ourthink}[1]{\textcolor{oursgreen}{#1}}
\newcolumntype{L}{>{\raggedright\arraybackslash}X}
\usepackage[T1]{fontenc}

\usepackage[utf8]{inputenc}

\usepackage{microtype}

\usepackage{inconsolata}

\usepackage{graphicx}

\title{CORA: Analyzing and bridging thinking-answer gap in Multimodal RLVR via Consistency-Oriented Reasoning Alignment}

\author{
  \textbf{Jiayue Cao\textsuperscript{1*}},
  \textbf{Zhicong Lu\textsuperscript{1*}},
  \textbf{Xuehan Sun\textsuperscript{2}},
  \textbf{Wei Jia\textsuperscript{1$\dagger$}},
  \textbf{Hongling Zheng\textsuperscript{2}}
\\
  \textbf{Changyuan Tian\textsuperscript{1}},
  \textbf{Zichuan Lin\textsuperscript{3}},
  \textbf{Wenqian Lv\textsuperscript{1}},
  \textbf{Nayu Liu\textsuperscript{4}}
\\
\\
  \textsuperscript{1}University of Chinese Academy of Sciences \\
  \textsuperscript{2}Wuhan University,
  \textsuperscript{3}Tsinghua University,
  \textsuperscript{4}Tianjin University
}

\begin{document}
\maketitle

\begingroup
\renewcommand{\thefootnote}{\fnsymbol{footnote}}
\footnotetext[1]{Equal contribution.}
\footnotetext[2]{Corresponding author.}
\endgroup

\begin{abstract}

Reinforcement learning with verifiable rewards (RLVR) has successfully elicited the reasoning capabilities of large language models, motivating its extension to multimodal scenarios.
Existing methods primarily focus on improving the visual coverage of reasoning traces and mitigating visual hallucinations, but underestimate the semantic inconsistency between the reasoning process and the final answer. In this paper, we delve into thinking-answer inconsistency in RLVR for large vision-language models (LVLMs), showing thorough analyses of rollouts collected throughout Group Relative Policy Optimization (GRPO) training process and post-RLVR evaluation outputs that this issue persists during training and remains present during inference.
Motivated by the analysis, we propose \textbf{C}onsistency-\textbf{O}riented \textbf{R}easoning \textbf{A}lignment (\textbf{CORA}), which introduces thinking-answer semantic consistency into RLVR through a lightweight plug-and-play consistency reward model, and further incorporates Hybrid Reward Advantage Splitting (HRAS) to stably coordinate task and consistency optimization.
Extensive experiments across representative multimodal reasoning benchmarks and mainstream LVLMs show that CORA improves task performance while effectively mitigating thinking-answer inconsistency, leading to more faithful reasoning traces.

\end{abstract}

\section{Introduction}

Reinforcement learning with verifiable rewards (RLVR) has recently shown strong effectiveness in enhancing the reasoning capabilities of large language models~\citep{guo2025deepseek,  chen2026towards, jia2025u, diao2026hipif, lu2025piper}. Building on this success, a growing line of work seeks to extend RLVR to large vision-language models (LVLMs), aiming to enhance their ability to perform complex multimodal reasoning~\citep{liu2025visual, huang2025vision,feng2026video}.

\begin{figure}[t]
  \centering
  \includegraphics[width=0.99\linewidth]{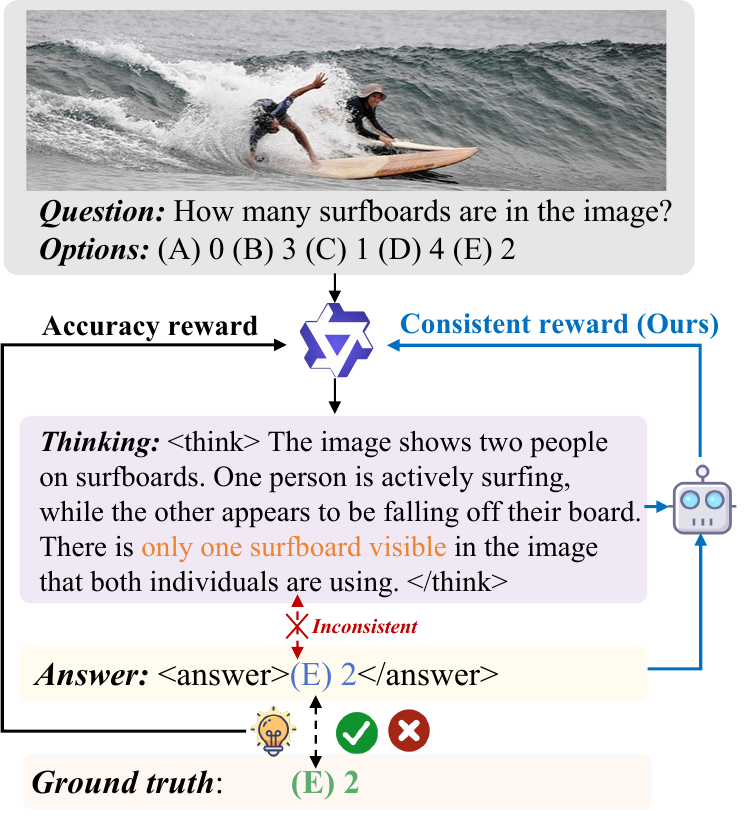}
  \caption{Thinking-answer inconsistency in multimodal RLVR. Existing works typically use final-answer correctness as the accuracy reward, underestimating potential inconsistencies between the reasoning trace and the final answer. Our method introduces the consistency reward to encourage the model to derive final answers from faithful reasoning traces.}
  \label{fig:intro_case}
\end{figure}

Following the standard RLVR paradigm that generates explicit reasoning traces before final answers, previous methods~\citep{man2025argus, lim2026vg} primarily focus on improving visual coverages and reducing hallucinations of reasoning traces. However, they rely on a foundational assumption: reasoning traces foster faithful final-answer generation, which can be fragile under answer-level rewards, where reasoning and answer may diverge despite the answer being correct.
Concurrent works~\citep{grpo-care,huang2025answer} either heuristically recognize this reasoning-answer mismatch within specific task settings or rely on costly expensive auxiliary reward mechanisms, leaving the dynamic evolution of such inconsistency during training insufficiently explored.

To further investigate this issue, we conduct an empirical analysis of on-policy rollouts during  group relative policy optimization (GRPO)~\citep{shao2024deepseekmath, lu2026hisr} training and evaluation. Specifically, we analyze serval mainstream open-source LVLMs~\citep{wang2024qwen2,qwen2.5vl} of different scales throughout the training process, thereby tracing how thinking-answer inconsistency evolves dynamically during RLVR. To ensure the comprehensive analysis, we select common scenarios~\citep{ray2024sat, shi2024math, lu2024mathvista,ghosal2025algopuzzlevqa} in multimodal reasoning, including spatial reasoning, multimodal mathematical reasoning, and puzzle. We observe that thinking-answer inconsistency is not merely caused by a small number of outlier cases, but widely emerges during RLVR with GRPO. Moreover, this issue consistently persists throughout the training dynamics, rather than being naturally mitigated as RLVR training progresses. In addition, standard GRPO is insufficient to reliably mitigate this issue as training progresses. Even after RLVR, models can still produce reasoning traces that fail to support the final answer or even semantically contradict it during evaluation. We attribute this issue arises from the answer-level reward design, which supervises only the final answer. This may lead the model to learn shortcuts for reaching correct answers, rather than deriving them from the generated reasoning traces.

Motivated by above evaluation observations and analyses, we propose \textbf{C}onsistency-\textbf{O}riented \textbf{R}easoning \textbf{A}lignment (\textbf{CORA}), which explicitly regularizes the semantic consistency between the thinking process and the answer during RLVR. Specifically, we introduce a lightweight and plug-and-play Consistency Reward Model (CRM) that scores thinking-answer consistency in a Natural Language Inference (NLI)-style discriminative manner, and incorporate this signal into GRPO as an additional reward. Moreover, to properly incorporate a continuous consistency reward while avoiding conflicts with the original discrete accuracy reward under group-wise normalization, we propose Hybrid Reward Advantage Splitting (HRAS), a reward-decoupled advantage estimation strategy that preserves the distinct preference signals of task and consistency rewards through separate group-wise normalization and weighted advantage composition.

To validate the effectiveness and generalization of our method, we carry out extensive experiments on three scenarios of multimodal reasoning benchmarks with mainstream LVLMs. Compared with RLVR using standard GRPO, our method achieves stronger performance and effectively mitigates the semantic inconsistency between thinking and answers, further demonstrating its effectiveness and generalizability. Overall, our contributions are summarized as follows:

1) We conduct an in-depth analysis of thinking-answer inconsistency in RLVR for multimodal reasoning, showing that this issue persists throughout GRPO training and is not naturally mitigated in post-RLVR evaluation.

2) We propose CORA, a consistency-oriented RLVR method that introduces a lightweight and plug-and-play consistency reward model to enhance thinking-answer semantic consistency, together with HRAS for stable joint optimization of task and consistency rewards.

3) Extensive experiments on multiple representative multimodal reasoning benchmarks demonstrate that CORA consistently improves performance across mainstream LVLMs while effectively mitigating thinking-answer inconsistency.
The code will be released soon to foster future research in RLVR for multimodal reasoning.

\section{Thinking-Answer Inconsistency in RLVR}
In this section, we first define a binary consistency measure to evaluate the degree of consistency between thinking and answer during RLVR-based training and inference. We then conduct a series of exploratory experiments~\citep{lu2024rethinking} to investigate whether LVLMs trained with answer-correctness rewards exhibit inconsistency between their thinking process and final answer. Finally, we summarize the quantitative findings and provide qualitative analyses of this issue.

\subsection{Thinking-Answer Consistency Measure}
Given a question $q$, a reasoning process $t$ within the \texttt{<think>} tag, and a final answer $a$ within the \texttt{<answer>} tag, we define thinking-answer consistency as a binary judgment. A sample is labeled as \textit{consistent} if the thinking $t$ semantically supports or leads to the answer $a$; otherwise, it is labeled as \textit{inconsistent}. Inconsistent cases include: (1) the answer implied by the thinking contradicts the final answer; and (2) the reasoning process fails to provide substantive evidence for the final answer. We use the Inconsistency Rate (IR) to quantify the degree of thinking-answer inconsistency during training and inference.

\begin{equation}
\mathrm{IR} = \frac{N_{\mathrm{incons}}}{N_{\mathrm{valid}}},
\label{eq:inconsistency-rate}
\end{equation}
where $N_{\mathrm{incons}}$ denotes the number of samples labeled as inconsistent, and $N_{\mathrm{valid}}$ denotes the total number of valid samples whose thinking and answer can be extracted and evaluated.To investigate whether the generated thinking reliably supports the final answer during RLVR-based training and inference, we conduct the following exploratory experiments.

\subsection{Empirical Analysis}
\label{sec:diagnosis}
\begin{figure}[t]
  \centering
  \includegraphics[width=\columnwidth]{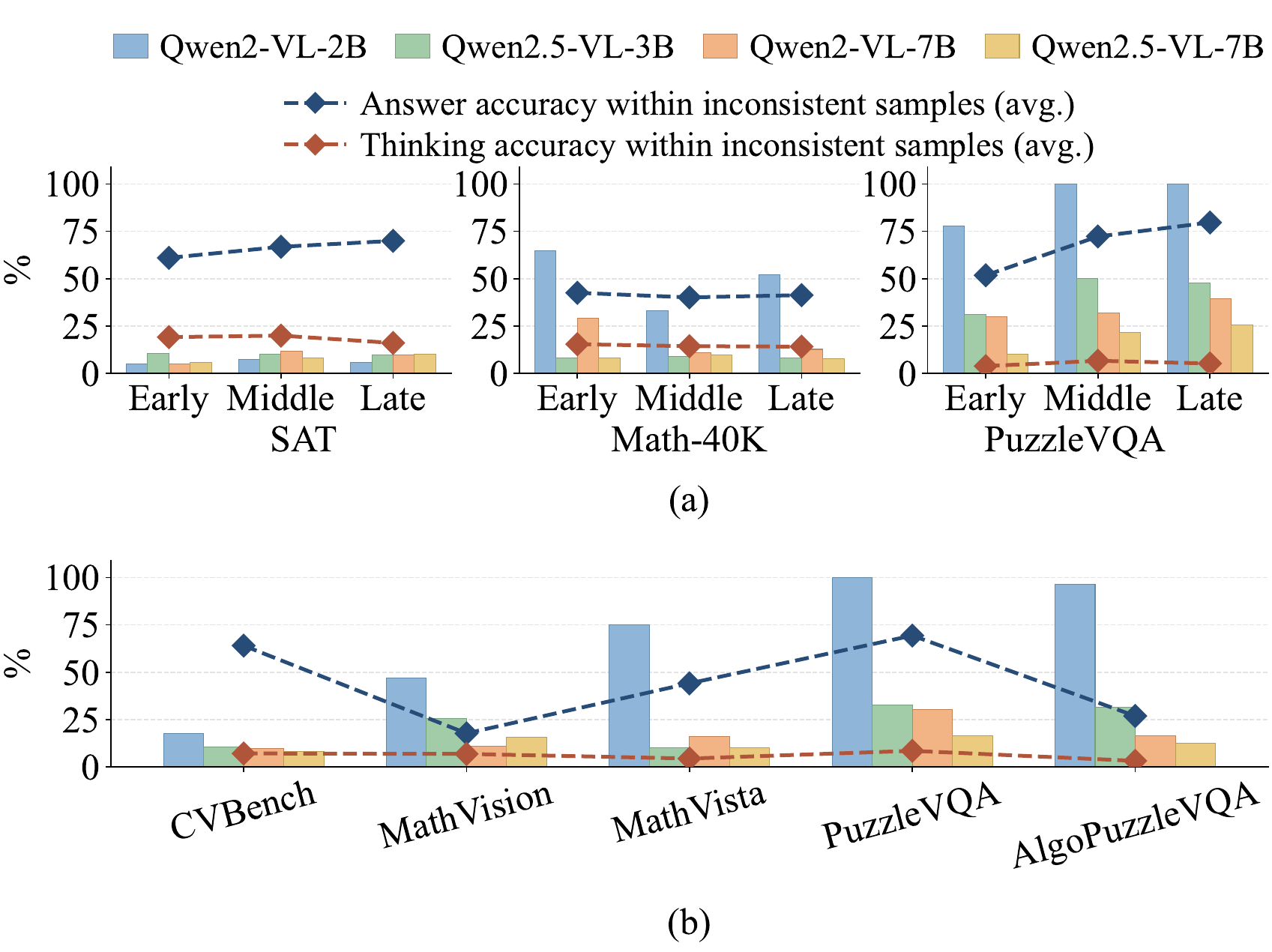}
  \caption{Comparison of inconsistency rate and thinking-vs-answer accuracy among inconsistent samples. (a) Phenomenon during training. (b) Phenomenon on evaluation benchmarks.}
  \label{fig:inconsistency-compact}
\end{figure}

\textbf{Data.}
We focus on three representative multimodal reasoning tasks: visual perception, including classification and spatial grounding; multimodal mathematical reasoning; and visual puzzle reasoning. For visual perception, we train on the SAT dataset~\citep{ray2024sat} and evaluate on CVBench dataset~\citep{tong2024cambrian}. For multimodal mathematical reasoning, we train on Math-40K~\citep{shi2024math} and evaluate on MathVision~\citep{wang2024measuring} and MathVista~\citep{lu2024mathvista}. For visual puzzle reasoning, we follow the setting of~\citep{li2025think} and construct a 6.5K-sample PuzzleVQA training set using the generation code released by~\citep{chia2024puzzlevqa}. During evaluation, we use PuzzleVQA as the in-domain test set and AlgoPuzzleVQA~\citep{ghosal2025algopuzzlevqa} as the out-of-domain test set.

\noindent
\textbf{Experimental Setup.}
To investigate whether thinking and answers remain consistent in multimodal reasoning, we conduct a preliminary study under a standard RLVR setting. We train four Qwen-series LVLMs on the three representative multimodal reasoning tasks described above, using format compliance and answer correctness as reward signals. The experimental details are kept consistent with the main experiments and are provided in Section~\ref{experimental_setup}.

After training, we construct two corpora for consistency analysis. For the training-stage corpus, model states from the early, middle, and late stages of each training run are selected and used to generate responses on the corresponding training prompts. We then extract the thinking and answer contents from each response, resulting in a training-stage thinking-answer corpus with approximately 719K valid samples. For the evaluation-stage corpus, the final trained model from each run is used to generate responses on evaluation benchmarks, and the corresponding thinking-answer pairs are extracted in the same manner.

To annotate sample-level consistency in the corpus, we design a dedicated judging prompt, which is provided in the Appendix~\ref{sec:appendix-prompts}. The prompt instructs the judge model to perform three steps for each response: (1) extract the final conclusion from the reasoning traces within the \texttt{<think>} tag; (2) extract the final answer from the \texttt{<answer>} tag; and (3) determine whether the thinking process and the final answer point to the same semantic answer. The annotation rules are as follows: if the conclusion from the thinking process is semantically equivalent to the final answer, the consistency label is \textit{Yes}; if they point to different values, the label is \textit{No}; and if either the reasoning process or the final answer does not contain an extractable answer, the label is \textit{None}. The first case is treated as consistent, while the latter two cases are treated as inconsistent. Across all datasets, a unified annotation template is adopted, with dataset-specific few-shot examples provided to accommodate different task formats.

During annotation, GPT-5~\citep{openai2025gpt5systemcard} is used as the judge model. Based on these annotations, we quantify whether the thinking process and final answer remain consistent during RLVR training and final inference, as well as the extent of inconsistency when it occurs.

\begin{figure*}[t]
  \centering
  \includegraphics[width=0.99\textwidth]{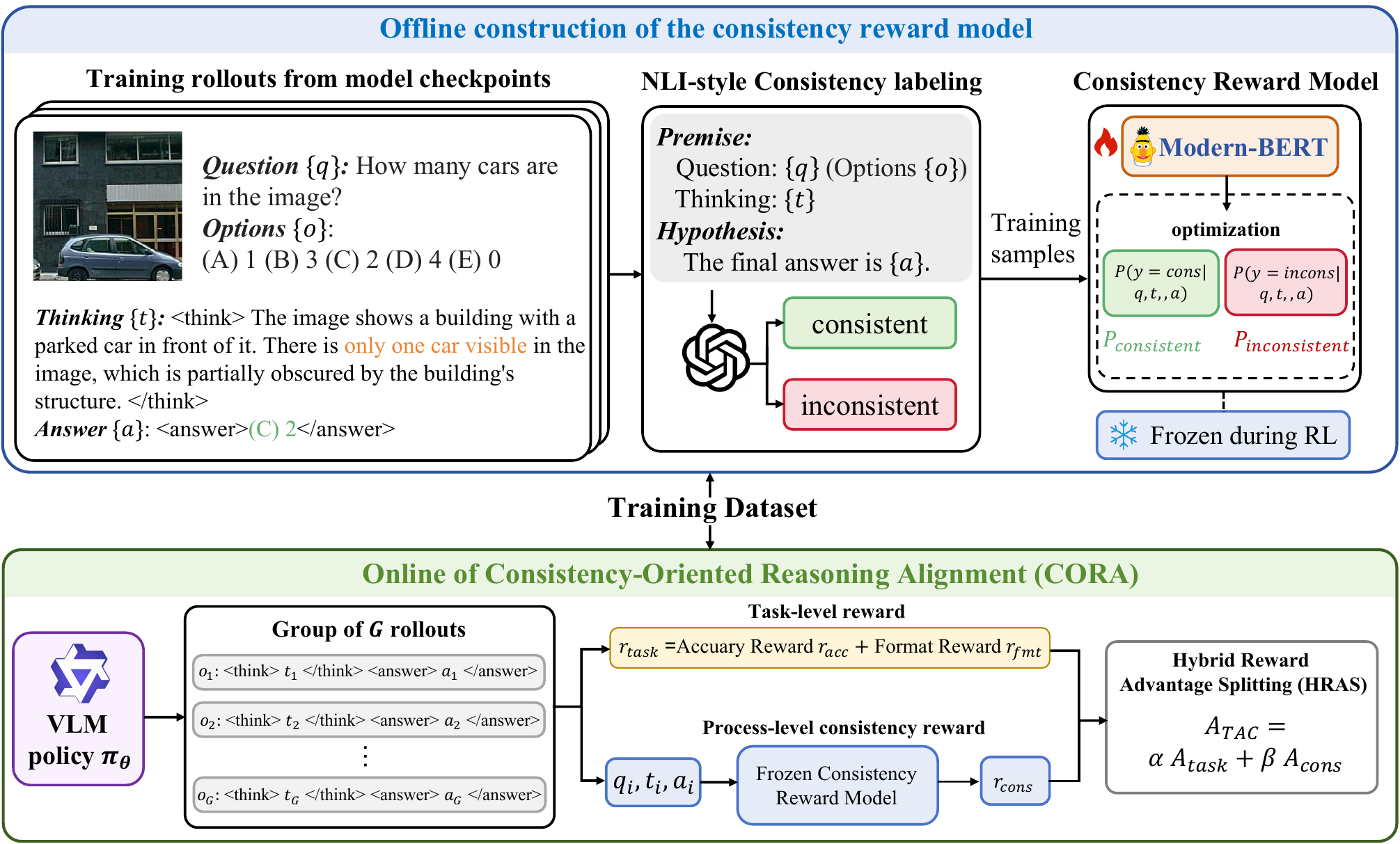}
  \caption{Overview of CORA. We construct an NLI-style consistency discrimination dataset from the training-stage corpus to train a ModernBERT-based CRM. During GRPO, the CRM provides a consistency reward for each generated \texttt{<think>}--\texttt{<answer>} pair. CORA separately normalizes task and consistency rewards into split advantages and combines them for policy optimization.}
  \label{fig:method-overview}
\end{figure*}

\subsection{Quantitative Results and Analysis}

Figure~\ref{fig:inconsistency-compact} compares thinking-answer inconsistency during training and evaluation. The results show that inconsistency exists in both stages of LVLM behavior and is not an occasional failure case, but a widespread phenomenon across models and visual reasoning tasks. The inconsistency rate is task-dependent: it is lower on visual perception tasks, where answers can often be directly inferred from visual evidence, but higher on multimodal mathematical reasoning and visual puzzle reasoning tasks, which require deeper computation and multi-step inference. Moreover, inconsistency persists throughout training and often becomes more pronounced as training proceeds. The same issue further appears on evaluation benchmarks. Finally, among inconsistent samples, the final answer is more often correct than the reasoning-implied answer. This suggests that current RLVR training primarily optimizes answer correctness while leaving the thinking process under-supervised.

\section{Method}

In this section, we present the proposed CORA method. As illustrated in Figure~\ref{fig:method-overview}, CORA employs the CRM to estimate thinking-answer consistency and integrates it into RLVR as an explicit reward signal. To mitigate interference among heterogeneous rewards, we introduce a reward-advantage splitting strategy that separates the optimization signals from task rewards and consistency rewards.

\subsection{Consistency Reward Model}
Based on the preliminary analysis, we argue that RLVR with answer-level rewards suffers from a lack of direct supervision over the thinking process. Although such rewards can encourage models to produce correct final answers, they do not explicitly verify whether the reasoning process semantically supports the final output answer. To generate more reliable and consistent reasoning, we introduce an explicit consistency reward into the RLVR process.

Specifically, we formulate the problem of judging thinking-answer consistency as a NLI task. We then construct a labeled dataset and train a lightweight consistency reward model to score, during rollout, whether the generated thinking process is semantically consistent with the final answer. This score is incorporated into training as an additional consistency reward.

\textbf{NLI-style formulation.}
Given a question $q$, thinking process $t$, and answer $a$, we formulate thinking-answer consistency as a NLI problem. Specifically, we combine $q$ and $t$ as the premise, and treat $a$ as the hypothesis.
\begin{equation}
\begin{aligned}
P(q,t) &= [\textrm{Question: } q;\ \textrm{Thinking: } t], \\
H(a) &= [\textrm{The final answer is: } a].
\end{aligned}
\label{eq:nli-input}
\end{equation}

When $P(q,t)$ semantically entails $H(a)$, the thinking and the answer are considered consistent; otherwise, they are considered inconsistent.

\textbf{Dataset construction.}
We construct the dataset for training the CRM using the training-stage corpus described in Section~\ref{sec:diagnosis}. For each sample in the corpus, we extract the question, thinking process, and final answer, and then convert them into NLI-style premise-hypothesis pairs following the format in Eq.~\ref{eq:nli-input}. We further perform stratified sampling according to data source, thinking-answer consistency label, and training stage. Finally, we obtain a consistency discrimination dataset containing approximately 90K samples. The dataset is split into training and test sets with an 8:2 ratio. The training set is used to train the CRM, while the test set is used to evaluate its ability to distinguish thinking-answer consistency.

\textbf{NLI-based Consistency Reward Model.}
During RLVR training, the CRM needs to be invoked for every rollout to produce a consistency reward, making both inference efficiency and scoring quality important. We adopt ModernBERT~\citep{warner2025smarter} as a lightweight discriminative encoder for consistency reward prediction. ModernBERT supports long-context inputs of up to 8192 tokens and is well suited for classification tasks, making it appropriate for our NLI-based consistency scoring setting. Compared with using a LLM as an online judge during training, this design provides a more efficient and cost-effective reward function.

We fine-tune ModernBERT on the constructed 90K NLI-style consistency discrimination dataset using a cross-entropy loss. During training, the premise and hypothesis are jointly encoded. The hidden representation of the \texttt{[CLS]} token is then fed into a two-layer classification head to predict a binary label, i.e., consistent or inconsistent. Details of the CRM training hyperparameters and evaluation results are provided in Appendix~\ref{sec:appendix-reward-model}.

After training, the CRM is frozen and used as a scoring function during the RLVR stage. The weighted consistency reward is then defined as
\begin{equation}
r_{\mathrm{cons}}(q,t,a)
= \lambda_{\mathrm{cons}} \cdot p_\phi(y=\mathrm{cons}\mid q,t,a),
\label{eq:cons-reward}
\end{equation}
where $\lambda_{\mathrm{cons}}$ controls the contribution of the consistency reward and $y$ denotes the thinking-answer consistency label.

\subsection{Hybrid Reward Advantage Splitting}

\begin{table*}[t]
\centering
\normalsize
\setlength{\tabcolsep}{5pt}
\renewcommand{\arraystretch}{1.05}
\resizebox{0.98\textwidth}{!}{%
\begin{tabular}{l|l|cc|cc|cc|cc|cc}
\toprule
\multirow{2}{*}{Model} & \multirow{2}{*}{Method}
  & \multicolumn{2}{c|}{CVBench}
  & \multicolumn{2}{c|}{MathVision}
  & \multicolumn{2}{c|}{MathVista}
  & \multicolumn{2}{c|}{PuzzleVQA}
  & \multicolumn{2}{c}{AlgoPuzzleVQA} \\
\cmidrule(lr){3-4}
\cmidrule(lr){5-6}
\cmidrule(lr){7-8}
\cmidrule(lr){9-10}
\cmidrule(lr){11-12}
& &
  Acc$\uparrow$ & IR$\downarrow$
  & Acc$\uparrow$ & IR$\downarrow$
  & Acc$\uparrow$ & IR$\downarrow$
  & Acc$\uparrow$ & IR$\downarrow$
  & Acc$\uparrow$ & IR$\downarrow$ \\
\midrule

\multirow{2}{*}{Qwen2-VL-2B}
  & Baseline
  & 68.12 & 17.77
  & \textbf{15.36} & \textbf{46.96}
  & 46.10 & 74.90
  & 51.10 & 100
  & \textbf{29.39} & \textbf{96.55} \\
  & CORA
  & \textbf{68.39} & \textbf{12.44}
  & 11.71 & 54.78
  & \textbf{47.00} & \textbf{55.46}
  & \textbf{63.35} & \textbf{99.90}
  & 28.28 & 99.09 \\
\midrule

\multirow{2}{*}{Qwen2-VL-7B}
  & Baseline
  & 77.52 & 9.56
  & 17.53 & 10.76
  & \textbf{58.70} & 16.09
  & 77.00 & 30.21
  & \textbf{26.72} & 16.46 \\
  & CORA
  & \textbf{78.09} & \textbf{8.64}
  & \textbf{18.78} & \textbf{9.71}
  & 58.50 & \textbf{15.26}
  & \textbf{81.95} & \textbf{5.03}
  & 26.67 & \textbf{5.31} \\
\midrule

\multirow{2}{*}{Qwen2.5-VL-3B}
  & Baseline
  & 76.04 & 10.43
  & 19.01 & 25.52
  & 59.40 & \textbf{10.05}
  & \textbf{65.45} & 32.71
  & 28.78 & 31.54 \\
  & CORA
  & \textbf{76.50} & \textbf{9.11}
  & \textbf{19.34} & \textbf{24.95}
  & \textbf{61.30} & 15.08
  & 64.35 & \textbf{24.06}
  & \textbf{30.17} & \textbf{30.38} \\
\midrule

\multirow{2}{*}{Qwen2.5-VL-7B}
  & Baseline
  & 77.75 & 8.08
  & 12.99 & 15.80
  & 67.90 & 9.89
  & 76.10 & 16.29
  & 31.28 & 12.54 \\
  & CORA
  & \textbf{78.24} & \textbf{7.36}
  & \textbf{17.00} & \textbf{8.28}
  & \textbf{69.30} & \textbf{6.48}
  & \textbf{76.60} & \textbf{3.37}
  & \textbf{31.83} & \textbf{2.50} \\
\bottomrule
\end{tabular}%
}
\caption{Main results on five multimodal reasoning benchmarks across four Qwen-VL backbones. Best results are highlighted in \textbf{bold}.}
\label{tab:main}
\end{table*}


Standard GRPO samples a group of responses for each prompt and sums all reward components of each response into a single scalar reward. The advantage is then computed by normalizing these summed rewards within the response group. However, different rewards may have different distributional properties, and directly summing them before advantage computation can introduce scale coupling among heterogeneous reward signals. Specifically, format rewards and answer-correctness rewards are typically sparse and discrete, whereas the consistency rewards produced by the CRM are continuous probability scores. When rewards with different distributions are directly added and normalized together, the reward component with larger within-group variance may dominate the final advantage estimate, thereby weakening the advantage signals from other rewards. For consistency-aware RLVR, this leads to a potential issue: if the consistency reward dominates, the model may overemphasize thinking-answer alignment while under-optimizing answer correctness. Conversely, if the answer-correctness reward dominates, the consistency signal may become too weak to effectively constrain the thinking process.

To address this issue, we propose HRAS. HRAS groups the format reward and answer-correctness reward as the task reward, and computes the advantages of the task reward and the consistency reward separately within each response group. The two advantages are then combined with weighted coefficients to form the final advantage, preventing interference between heterogeneous reward signals.

\begin{equation}
A_{\mathrm{TAC}}^{(i)}
=
\alpha
\frac{r_{\mathrm{task}}^{(i)}-\mu_{\mathrm{task}}}
{\sigma_{\mathrm{task}}+\epsilon}
+
\beta
\frac{r_{\mathrm{cons}}^{(i)}-\mu_{\mathrm{cons}}}
{\sigma_{\mathrm{cons}}+\epsilon}.
\label{eq:split-advantage}
\end{equation}
Here, $\mu_{\mathrm{task}}, \sigma_{\mathrm{task}}$ and $\mu_{\mathrm{cons}}, \sigma_{\mathrm{cons}}$ are computed separately over the $G$ responses sampled for the same prompt, and $\alpha,\beta$ control the relative strength of task learning and consistency learning.

\section{Experiments}

\subsection{Experimental Setup}
\label{experimental_setup}

\textbf{Dataset.} To evaluate the effectiveness of our method, we conduct RLVR experiments using GRPO on three scenarios of multimodal reasoning benchmarks, including CVBench, MathVision, MathVista, PuzzleVQA, and AlgoPuzzleVQA, following the setup in Section~\ref{sec:diagnosis}. More details of datasets are provided in Appendix~\ref{sec:appendix-implementation}.

\noindent
\textbf{Implementations.}
We use Qwen2-VL-2B/7B-Instruction and Qwen2.5-VL-3B/7B-Instruction as the backbone LVLMs. We perform RLVR on the training datasets from the three multimodal scenarios, respectively. The baseline follows the standard RLVR setting and uses only answer-correctness and format rewards, without the proposed consistency reward. CORA keeps the same task rewards and additionally incorporates the CRM-based consistency reward with HRAS. Across all experimental settings, we set the batch size to 1, use gradient accumulation of 2 steps, and set the format reward weight to 0.1. Training is conducted for two epochs on SAT and PuzzleVQA, and for one epoch on Math-40K. Other detailed training configurations are provided in Appendix~\ref{sec:appendix-implementation}.

\subsection{Main Result}

\begin{table*}[t]
\centering
\small
\setlength{\tabcolsep}{8pt}
\renewcommand{\arraystretch}{1.05}
\resizebox{0.99\textwidth}{!}{%
\begin{tabularx}{\textwidth}{l l CCCCC}
\toprule
Method & Metric & CVBench & MathVision & MathVista & PuzzleVQA & AlgoPuzzleVQA \\
\midrule
\multirow{3}{*}{CORA}
  & Acc $\uparrow$        & \textbf{78.24} & \textbf{17.00} & \textbf{69.3} & \textbf{76.6} & \textbf{31.83} \\
  & IR $\downarrow$   & 7.36 & \textbf{8.28} & 6.48 & 3.37 & \textbf{2.50} \\
  & ConsAcc $\uparrow$    & 80.03 & 26.38 & 73.24 & \textbf{76.56} & \textbf{32.68} \\
  \midrule
\multirow{3}{*}{w/o CRM}
  & Acc $\uparrow$        & 77.75 & 12.99 & 67.90 & 76.10 & 31.28 \\
  & IR $\downarrow$   & 8.08 & 15.80 & 9.89 & 16.29 & 12.54 \\
  & ConsAcc $\uparrow$    & 80.45 & \textbf{27.22} & \textbf{74.77} & 74.91 & 32.17 \\
\midrule
\multirow{3}{*}{w/o HRAS}
  & Acc $\uparrow$        & 76.12 & 14.08 & 68.1 & 75.3 & 30.78 \\
  & IR $\downarrow$   & \textbf{1.22} & 10.60 & \textbf{0.50} & \textbf{0.55} & 3.27 \\
  & ConsAcc $\uparrow$    & \textbf{80.49} & 23.90 & 70.49 & 75.57 & 30.90 \\

\bottomrule
\end{tabularx}
}
\caption{Ablation results of CRM and HRAS on Qwen2.5-VL-7B. ConsAcc denotes the answer accuracy among samples whose thinking process is consistent with the final answer. Best results are highlighted in \textbf{bold}.}
\label{tab:ablation-splitadv}
\end{table*}

Table~\ref{tab:main} reports the accuracy and inconsistency rate on five evaluation benchmarks across three multimodal reasoning tasks. Overall, in most settings, CORA substantially reduces the inconsistency between thinking and answer while also improving final-answer accuracy. We further observe that the performance gains brought by CORA are more pronounced on 7B-scale backbones than on smaller 2B/3B models. A possible explanation is that consistency supervision assumes that the model can first generate a meaningful reasoning process. Larger models typically possess stronger reasoning capabilities, making their generated thinking content more amenable to consistency evaluation and optimization. By contrast, smaller models may generate incomplete or vacuous reasoning traces, thereby limiting the effectiveness of the consistency reward mechanism.

Moreover, the effectiveness of CORA is also task-dependent. The performance gains are more pronounced on MathVision and AlgoPuzzleVQA. This is because both benchmarks require deeper reasoning capabilities, where the final answer depends heavily on the intermediate reasoning process. CORA provides an explicit process-level signal that encourages the generated thinking process to more effectively support the final answer. The reward curves in Appendix~\ref{sec:appendix-accuracy-and-consistency-reward-curves} support our claim.

Furthermore, we observe that reductions in inconsistency rate are associated with improvements in final-answer accuracy. This suggests that thinking-answer consistency is closely related to the model's task-solving ability. When the generated thinking semantically supports the final answer, the final prediction is more likely to arise from a coherent and effective reasoning process. This observation further supports our hypothesis that RLVR training based solely on final-answer correctness may not sufficiently supervise the thinking process, whereas explicitly introducing a consistency reward can encourage the model to generate more reliable reasoning and thereby improve final-answer correctness.

\subsection{Ablation Study}

To verify the effect of the CRM and HRAS, we perform ablation studies using Qwen2.5-VL-7B as the backbone. When ablating HRAS, we keep the task reward and consistency reward formulations unchanged, and replace Eq.~\ref{eq:split-advantage} with a conventional single-advantage computation based on the summed reward. 

As shown by the answer accuracy results in Table~\ref{tab:ablation-splitadv}, the proposed CORA method achieves the best performance across all five evaluation benchmarks. The w/o CRM and w/o HRAS variants exhibit mixed gains and drops across different benchmarks, but both underperform CORA overall. This suggests that CORA effectively balances reasoning consistency with answer correctness.

Directly adding the consistency reward to the total reward can achieve stronger inconsistency reduction on some benchmarks. However, this variant consistently underperforms CORA in final-answer accuracy. We attribute this trade-off to the training dynamics of merged reward advantage computation. In the later stage of training, responses within the same GRPO group often show limited variation in answer correctness, while their consistency scores may still vary substantially. As a result, the advantage signal produced by merged reward computation can be dominated by the consistency signal. This tendency may drive policy updates away from optimizing answer correctness. The fact that CORA achieves higher ConsAcc than the w/o HRAS variant on most benchmarks provides further supporting evidence.

\subsection{Analyses}
\begin{table}[t]
\centering
\setlength{\tabcolsep}{3.8pt}
\renewcommand{\arraystretch}{1.1}
\small
\resizebox{0.48\textwidth}{!}{%
\begin{tabular}{llcccc}
\toprule
\textbf{Dataset} & \textbf{Method} & \textbf{CC} & \textbf{CW} & \textbf{IC} & \textbf{IW} \\
\midrule
\multirow{3}{*}{MathVision}
 & Baseline   & 9.08  & 30.20 & 1.48  & 59.24 \\
 & CORA       & \textbf{15.46} & \textbf{53.75} & \textbf{1.02}  & \textbf{29.77} \\
 & $\Delta$ pp & +6.38 & +23.55 & $-$0.46 & $-$29.47 \\
\midrule
\multirow{3}{*}{PuzzleVQA}
 & Baseline   & 62.55 & 20.95 & 12.90 & 3.60 \\
 & CORA       & \textbf{73.55} & \textbf{22.65} & \textbf{2.85}  & \textbf{0.95} \\
 & $\Delta$ pp & +11.00 & +1.70 & $-$10.05 & $-$2.65 \\
\bottomrule
\end{tabular}
}
\caption{Comparison of sample categorization between the baseline and CORA on MathVision and PuzzleVQA, using Qwen2.5-VL-7B as the backbone. CC denotes Consistent-Correct, CW denotes Consistent-Wrong, IC denotes Inconsistent-Correct, and IW denotes Inconsistent-Wrong.}
\label{tab:four-quadrant}
\end{table}

\begin{figure*}[t]
  \centering
  \includegraphics[width=0.9\textwidth]{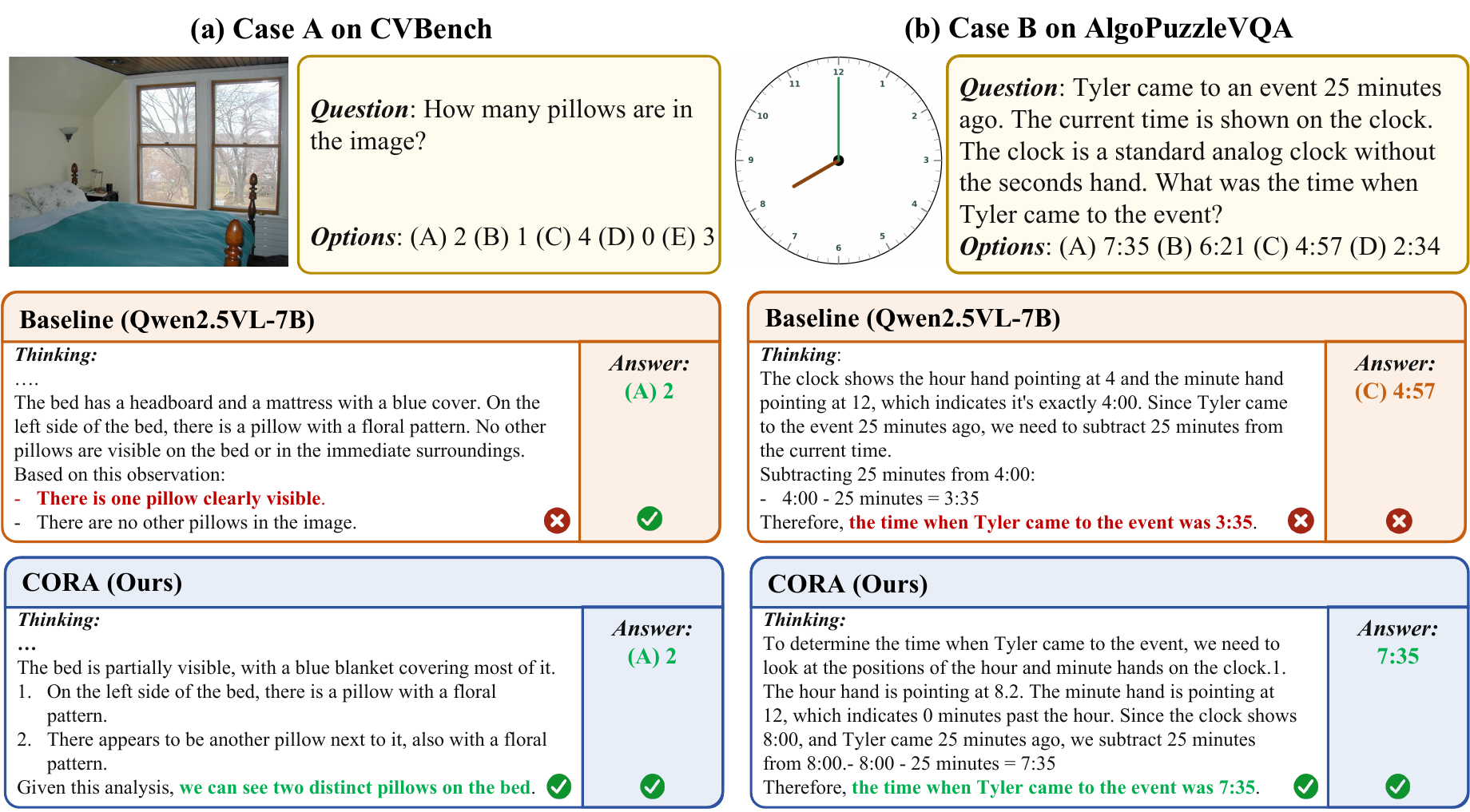}
  \caption{Case studies on different benchmarks.}
  \label{fig:case}
\end{figure*}

\textbf{Reliability Analysis.} 
To analyze whether CORA improves performance by promoting more reliable reasoning, we categorize model responses into four groups according to answer correctness and thinking-answer consistency, as shown in Table~\ref{tab:four-quadrant}. 
We compare the group distributions between the baseline and CORA on MathVision and PuzzleVQA. 
The results in Table~\ref{tab:four-quadrant} show that CORA increases the proportion of responses that are both correct and consistent, while reducing the proportion of correct but inconsistent responses. 
This suggests that CORA promotes semantically supported correctness, where correct final answers are accompanied by reasoning traces that provide valid justification.

\noindent
\textbf{Case study.} Figure~\ref{fig:case} illustrates how CORA improves the reliability of generated reasoning. The baseline may output a correct answer with an inconsistent reasoning trace, as in the pillow-counting example, or fail both reasoning and prediction, as in the clock example. In contrast, CORA produces reasoning processes that explicitly support the final answers. These cases qualitatively demonstrate that CORA encourages the model to generate answers with semantically aligned and better-justified reasoning traces.

\section{Related Works}

\textbf{Multimodal Chain-of-Thought Reasoning.} 
Multimodal chain-of-thought reasoning enables LVLMs to generate intermediate reasoning steps before producing final answers. Early studies such as Multimodal-CoT~\citep{zhang2023multimodal} show that generating intermediate rationales can improve performance on multimodal question answering tasks. Recent studies move beyond purely text-based reasoning traces and emphasize the critical role of visual evidence in the reasoning process. Argus~\citep{man2025argus} introduces a grounded chain-of-thought reasoning paradigm that assists reasoning through visual attention revisiting. VG-CoT~\citep{lim2026vg} explicitly links each reasoning step to visual evidence, further highlighting the importance of trustworthy visual reasoning. These methods often implicitly assume that the reasoning process can be reliably aligned with the final answer, and mainly focus on improving the reasoning process itself. 
However, in generated responses, the reasoning trajectory is not necessarily consistent with the final answer.

\noindent
\textbf{RLVR for Multimodal Reasoning.}
Reinforcement Learning with Verifiable Rewards (RLVR) has recently emerged as an effective paradigm for improving the reasoning ability of large language models. Inspired by this success, some studies~\citep{tian2026faithful, tian2026rectify} have extended RLVR to multimodal reasoning. Visual-RFT~\citep{liu2025visual} adapts reinforcement fine-tuning to VLMs by designing verifiable rewards for perception and grounding tasks. Subsequent works further explore RLVR through CoT supervised fine-tuning~\citep{tan2026reason}, multimodal cold-start data~\citep{huang2025vision}, and iterative SFT-RL self-improvement~\citep{deng2025openvlthinker}. Most existing RLVR methods still primarily optimize rewards defined at the task-output level. Recent studies have begun to recognize this limitation and introduce additional process-oriented signals. However, concurrent works either heavily rely on ground-truth information or are restricted to specific multiple-choice settings~\citep{kan2025taco,huang2025answer}, making them difficult to generalize to diverse multimodal reasoning scenarios.

\section{Conclusion}

In this paper, we delve into thinking-answer inconsistency in RLVR for LVLMs, where the generated reasoning trace may fail to support, or even semantically contradict to the final answer.
Through thorough analyses of on-policy rollouts during GRPO training and post-RLVR evaluation outputs, we show that this issue persistently emerges during training and is not naturally mitigated by standard GRPO. 
To address this problem, we propose CORA, a consistency-oriented RLVR method that introduces a lightweight and plug-and-play consistency reward model to promote thinking-answer consistency. We further design HRAS to decouple task and consistency rewards during advantage estimation, enabling stable joint optimization of answer correctness and reasoning reliability. 
Extensive experimental results across representative multimodal reasoning benchmarks demonstrate that CORA improves task performance while effectively reducing thinking-answer inconsistency.

\section*{Limitations}

Despite the impressive results of our method, we have to admit our work has the following limitations: (1) due to computational constraints, we only evaluate the effectiveness of the proposed method on models with 7B parameters or fewer. Its applicability to larger-scale models remains to be further validated. (2) Our current study focuses on image-text multimodal reasoning tasks, while higher-dimensional visual inputs such as videos are not considered. Evaluating the proposed method on broader and more diverse tasks will be an important direction for our future work.


\bibliography{ref}

\appendix

\section{Implementation Details}
\label{sec:appendix-implementation}
We train the consistency reward model with ModernBERT-large and implement CORA, both using bf16 precision. The rollout temperature is set to 1.0 for RL training. The standard GRPO baselines are reproduced on the same training data with answer-correctness and format rewards. 
All training runs are conducted on 16 NVIDIA A100 GPUs.

For spatial perception, we fine-tune on SAT for 2 epochs and evaluate on CVBench; the maximum prompt and response lengths are both set to 1024. For multimodal mathematical reasoning, we fine-tune on Math-40K for 1 epoch and evaluate on MathVista and MathVision; the maximum prompt and response lengths are set to 4096 and 512, respectively. 
For visual puzzle reasoning, we fine-tune for 2 epochs on the generated PuzzleVQA training set following the generation pipeline used by~\citet{li2025think}, and evaluate on PuzzleVQA and AlgoPuzzleVQA; the maximum prompt and response lengths are both set to 1024.

For the consistency reward, we adopt two types of scheduling strategies: warmup and decay. Warmup means that the consistency reward is temporarily disabled at the beginning of training, allowing the model to first learn basic task-solving behavior from the task reward. Decay means that, after the consistency reward is activated, its weight is gradually reduced as training proceeds, thereby preventing this signal from dominating the later training stage. The three datasets use different scheduling schemes according to their task characteristics. During optimization, the advantages of the task-reward branch and the consistency-reward branch are computed separately, where $\alpha$ and $\beta$ denote the weights of the two branches, respectively.

Regarding Qwen2.5-VL-7B, we adopt both warmup and decay on the SAT dataset. The consistency reward is disabled for the first 200 steps. After that, its initial weight is set to 0.3 and decays exponentially with a half-life of 40 steps. The minimum weight threshold is set to 0.05. Once the weight decays below this threshold, the consistency reward is no longer applied. For advantage computation, the task and consistency branches are weighted by $\alpha$ and $\beta$ with values of 1.0 and 0.05, respectively. On the Math-40K dataset, we only apply decay. The initial weight of the consistency reward is set to 1.0, the half-life is set to 183 steps, and the minimum weight threshold is set to 0.05. For advantage computation, the task and consistency branch weights are 1.0 and 0.5, respectively. On the PuzzleVQA dataset, we only apply warmup. Specifically, the consistency reward weight is set to 0 for the first 400 steps. After the warmup stage, it is fixed at 1.0 and kept unchanged throughout the remaining training process. For advantage computation, the task and consistency branch weights are 1.0 and 0.5, respectively.

\section{Consistency Reward Model Details}
\label{sec:appendix-reward-model}

We train a binary consistency reward model to judge whether a response's reasoning trace supports its final answer. The classifier uses \texttt{answerdotai/ModernBERT-large} as the backbone and predicts two labels: \textit{consistent} and \textit{inconsistent}. Table~\ref{tab:bert-training-config} summarizes the training configuration, and Table~\ref{tab:bert-eval-results} reports the held-out evaluation results used by our consistency reward.

\section{Comparison with GPT-4o-mini Reward Model}
\label{sec:appendix-gpt-rm}

To verify whether CORA's effectiveness depends on using a large language model as the online consistency reward, we conduct an additional comparison by replacing our lightweight CRM with GPT-4o-mini as the consistency reward model. We keep the RLVR training setting the same as CORA and run the experiment on Qwen2.5-VL-7B, using SAT for RLVR training and CVBench for evaluation. Due to the high computational and monetary cost of invoking the closed-source LLM as the reward model during RL rollouts, we conduct this comparison only on this representative setting.

\begin{table}[t]
\centering
\small
\begin{tabularx}{\linewidth}{@{}lX@{}}
\toprule
\textbf{Item} & \textbf{Setting} \\
\midrule
Base model & ModernBERT-large \\
Number of labels & 2 \\
Maximum sequence length & 2048 \\
Batch size & 16 \\
Epochs & 3 \\
Learning rate & \(2\times10^{-5}\) \\
\bottomrule
\end{tabularx}
\caption{Training configuration for the binary consistency reward model.}
\label{tab:bert-training-config}
\end{table}

\begin{table}[t]
\centering
\small
\begin{tabular}{@{}lc@{}}
\toprule
\textbf{Metric} & \textbf{Score} \\
\midrule
Overall accuracy & 0.8734 \\
Macro F1 & 0.8733 \\
Weighted F1 & 0.8733 \\
\bottomrule
\end{tabular}
\caption{Overall the test set of the consistency reward model.}
\label{tab:bert-overall-results}
\end{table}

\begin{table}[t]
\centering
\scriptsize
\setlength{\tabcolsep}{3.5pt}
\begin{tabular}{@{}lrrrrr@{}}
\toprule
\textbf{Source} & \textbf{\(n\)} & \textbf{Acc.} & \textbf{Macro F1} & \textbf{R(con)} & \textbf{R(inc)} \\
\midrule
Math & 10014 & 0.8097 & 0.8090 & 0.8613 & 0.7571 \\
Puzzle & 6037 & 0.9730 & 0.9730 & 0.9707 & 0.9753 \\
Sat & 2036 & 0.8919 & 0.8919 & 0.9108 & 0.8742 \\
\bottomrule
\end{tabular}
\caption{Consistency reward model performance by source dataset. R(con) and R(inc) denote recall for the consistent and inconsistent classes, respectively.}
\label{tab:bert-eval-results}
\end{table}

\begin{table}[!htbp]
\centering
\scriptsize
\setlength{\tabcolsep}{3pt}
\begin{tabularx}{\columnwidth}{@{}lXccc@{}}
\toprule
\textbf{Type} & \textbf{Metric} & \textbf{CORA} & \textbf{GPT-RM} & \textbf{\(\Delta\)} \\
\midrule
\multirow{2}{*}{Overall} 
& Accuracy \(\uparrow\) & 78.24\% & 75.32\% & +2.92 \\
& Inconsistency rate \(\downarrow\) & 7.36\% & 5.95\% & +1.41 \\
\midrule
\multirow{4}{*}{Detailed analysis}
& CC: consistent and correct \(\uparrow\) & 72.25\% & 71.38\% & +0.87 \\
& CW: consistent but wrong \(\downarrow\) & 18.95\% & 22.14\% & -3.18 \\
& IC: inconsistent but correct & 5.99\% & 3.94\% & +2.05 \\
& IW: inconsistent and wrong \(\downarrow\) & 2.81\% & 2.54\% & +0.27 \\
\bottomrule
\end{tabularx}
\caption{Comparison between CORA and the GPT-4o-mini reward model on CVBench after SAT training with Qwen2.5-VL-7B. GPT-RM denotes replacing our lightweight CRM with GPT-4o-mini as the consistency reward model. CC denotes consistent and correct, and CW denotes consistent but wrong. \(\Delta\) is CORA minus GPT-RM.}
\label{tab:gpt-rm-comparison}
\end{table}

As shown in Table~\ref{tab:gpt-rm-comparison}, GPT-RM obtains a lower inconsistency rate, but its answer accuracy is lower than CORA by 2.92 points. To further analyze the reason behind this gap, we use GPT-5 to perform the judgment of the consistency between each generated thinking trace and its final answer after inference, and divide the responses according to both consistency and answer correctness. 
The results show that GPT-RM has a larger fraction of consistent-but-wrong responses: GPT-RM produces 22.14\% CW samples, while CORA reduces this group to 18.95\%. Meanwhile, CORA produces more consistent-and-correct responses than GPT-RM. This indicates that the lower inconsistency rate of GPT-RM is partly caused by generating responses whose thinking and answer are consistently wrong, rather than by improving task correctness. The possible reason is that a closed-source LLM judge may rely on its own prior knowledge when scoring the rollout, which does not necessarily provide a stable reward signal for aligning the model's internal thinking with its answer. In contrast, our lightweight CRM is explicitly trained to estimate thinking-answer consistency and can be plugged into RLVR training after a single efficient training stage, providing a lower-cost and faster reward model while preserving stronger answer accuracy.

\section{Prompt Templates for Consistency Annotation}
\label{sec:appendix-prompts}

Tables~\ref{tab:prompt-sat-train}--\ref{tab:prompt-test} report the API prompts used for thinking-answer consistency annotation on the training-stage corpus and the evaluation-stage corpus.

\begin{table*}[t]
\centering
\small
\begin{tabularx}{\textwidth}{p{0.16\textwidth}L}
\toprule
\textbf{Usage} & SAT training-rollout consistency annotation \\
\midrule
\textbf{Prompt} &
I will give you a SAT (spatial reasoning) question, its options, and a model's response. The model response contains a thinking process in \texttt{<think>...</think>} tags and a final answer in \texttt{<answer>...</answer>} tags.

SAT questions usually involve counting objects, judging spatial relations (left/right, above/below, in-front/behind), estimating depth/distance from camera, or identifying which marked object is closer/farther.

Your task: (1) read the thinking process and determine the final conclusion the reasoning arrived at, focusing on the conclusion sentence(s); (2) read the \texttt{<answer>} tag and copy the value the model committed to; (3) judge whether the thinking conclusion and the answer tag are consistent.

Important rules: \texttt{think\_conclusion} is the final value the thinking process actually concludes with. Copy explicit final values even if they are outside the option set. For yes/no questions, infer yes/no if the reasoning implies an affirmative or negative conclusion. Pay attention to negation and perspective equivalence. Output \texttt{None} if the thinking is generic or lists candidates without choosing one. \texttt{think\_in\_options} is \texttt{Yes}, \texttt{No}, or \texttt{None}; \texttt{answer\_tag} copies the value inside \texttt{<answer>} even if outside the options; \texttt{answer\_in\_options} is \texttt{Yes}, \texttt{No}, or \texttt{None}. \texttt{consistency} is \texttt{Yes} if the two values refer to the same value after normalization, \texttt{No} if they differ, and \texttt{None} only when either side is missing.

Output exactly five lines: \texttt{think\_conclusion: <value, option, or None>}; \texttt{think\_in\_options: <Yes/No/None>}; \texttt{answer\_tag: <value, option, or None>}; \texttt{answer\_in\_options: <Yes/No/None>}; \texttt{consistency: <Yes/No/None>}.

\textit{Few-shot SAT examples are inserted here.}

Now analyze this sample: Question: \{question\}; Options: \{options\}; Model response: \{response\}. \\
\bottomrule
\end{tabularx}
\caption{Prompt template for annotating think-answer consistency in SAT training rollouts. The full script inserts SAT-specific few-shot examples covering counting, spatial relations, yes/no inference, perspective equivalence, negation, and answers outside the option set.}
\label{tab:prompt-sat-train}
\end{table*}

\begin{table*}[t]
\centering
\small
\begin{tabularx}{\textwidth}{p{0.16\textwidth}L}
\toprule
\textbf{Usage} & Math-40K training-rollout consistency annotation \\
\midrule
\textbf{Prompt} &
I will give you a math/VQA question, its question type, its options (if any), and a model's response. The model response contains a thinking process in \texttt{<think>...</think>} tags and a final answer in \texttt{<answer>...</answer>} tags.

Question types in this dataset: \texttt{letter\_choice}, where the question has explicit choices such as (A)--(D), and \texttt{open\_ended}, where there is no fixed option set. For \texttt{letter\_choice}, the reasoning may state the option text instead of the letter; treat these as the same option. For \texttt{open\_ended}, both thinking and answer can be arbitrary text or numbers.

Your task: (1) read the thinking process and determine the final conclusion; (2) read the \texttt{<answer>} tag and copy the committed value; (3) judge whether the thinking conclusion and answer tag are consistent.

Important rules: for \texttt{letter\_choice}, \texttt{think\_conclusion} can be a letter, option text, or a raw value outside the options; for \texttt{open\_ended}, copy the final number, color, or identification phrase. Infer implicit yes/no conclusions, handle negation, and output \texttt{None} when the reasoning has no concrete conclusion. For \texttt{letter\_choice}, \texttt{think\_in\_options} and \texttt{answer\_in\_options} are \texttt{Yes}, \texttt{No}, or \texttt{None}; for \texttt{open\_ended}, they are always \texttt{N/A}. \texttt{consistency} is \texttt{Yes} when both sides refer to the same value after normalization or semantic equivalence, \texttt{No} when they differ, and \texttt{None} only when either side is missing.

Output exactly five lines with no extra text: \texttt{think\_conclusion: <value or None>}; \texttt{think\_in\_options: <Yes/No/None/N/A>}; \texttt{answer\_tag: <value or None>}; \texttt{answer\_in\_options: <Yes/No/None/N/A>}; \texttt{consistency: <Yes/No/None>}.

\textit{Few-shot Math-40K examples are inserted here.}

Now analyze this sample: Question type: \{question\_type\}; Options: \{options\}; Question: \{question\}; Model response: \{response\}. \\
\bottomrule
\end{tabularx}
\caption{Prompt template for annotating think-answer consistency in Math-40K training rollouts. The prompt supports both multiple-choice and open-ended questions and keeps out-of-option conclusions as raw values.}
\label{tab:prompt-math-train}
\end{table*}

\begin{table*}[t]
\centering
\small
\begin{tabularx}{\textwidth}{p{0.16\textwidth}L}
\toprule
\textbf{Usage} & PuzzleVQA training-rollout consistency annotation \\
\midrule
\textbf{Prompt} &
I will give you a puzzle question, its options, and a model's response. The model response contains a thinking process in \texttt{<think>...</think>} tags and a final answer in \texttt{<answer>...</answer>} tags.

The puzzles ask the model to find a missing element (number, color, shape, or size) in a visual pattern. The thinking process is often short and may not actually reach a conclusion that names one of the options.

Your task: (1) read the thinking process and determine the final conclusion; (2) read the \texttt{<answer>} tag and copy the committed value; (3) judge whether the thinking conclusion and answer tag are consistent.

Important rules: \texttt{think\_conclusion} is the final value the thinking actually concludes with. Copy explicit values even if they are outside the option set; output \texttt{None} if the thinking is generic or only lists candidates. \texttt{think\_in\_options} is \texttt{Yes}, \texttt{No}, or \texttt{None}. \texttt{answer\_tag} copies the value inside \texttt{<answer>} as-is, even if it is outside the options; \texttt{answer\_in\_options} is \texttt{Yes}, \texttt{No}, or \texttt{None}. \texttt{consistency} is \texttt{Yes} if both sides refer to the same normalized value, including when both are outside the options; \texttt{No} if they refer to different values; and \texttt{None} only when either side is missing. Compound colors such as ``light orange'' and ``orange'' must be treated as different values.

Output exactly five lines with no extra text: \texttt{think\_conclusion: <value or None>}; \texttt{think\_in\_options: <Yes/No/None>}; \texttt{answer\_tag: <value or None>}; \texttt{answer\_in\_options: <Yes/No/None>}; \texttt{consistency: <Yes/No/None>}.

\textit{Few-shot PuzzleVQA examples are inserted here.}

Now analyze this sample: Question: \{question\}; Options: \{options\}; Model response: \{response\}. \\
\bottomrule
\end{tabularx}
\caption{Prompt template for annotating think-answer consistency in PuzzleVQA training rollouts. The prompt explicitly preserves answers outside the option set and distinguishes compound colors.}
\label{tab:prompt-puzzle-train}
\end{table*}

\begin{table*}[t]
\centering
\small
\begin{tabularx}{\textwidth}{p{0.16\textwidth}L}
\toprule
\textbf{Usage} & Evaluation-set consistency extraction \\
\midrule
\textbf{Prompt} &
I will give you a question and a model response. The model response is in the form \texttt{<think>...</think><answer>...</answer>}. Your job is to extract two things and judge their consistency, looking only at the literal content of \texttt{<think>...</think>} and \texttt{<answer>...</answer>}. Do not use outside knowledge to guess what the correct answer should be. Do not bias your extraction toward any expected answer; only report what the model literally wrote.

Specifically: (1) \texttt{answer in thinking tag}: extract what the thinking process commits to as a final choice. If it does not explicitly commit to one of the listed options, output \texttt{None}. (2) \texttt{answer in answer tag}: extract whatever letter/value is literally inside \texttt{<answer>...</answer>}. If it is a single letter, expand it to its full option text using the option list; if it is already in ``(X) value'' form, keep it. (3) \texttt{consistency with answer tag}: output \texttt{Yes} if the two extracted values refer to the same option/value, \texttt{No} if they differ, and \texttt{None} if the thinking answer is \texttt{None}.

Output exactly these three lines: \texttt{answer in thinking tag:}; \texttt{answer in answer tag:}; \texttt{consistency with answer tag: <Yes/No/None>}.

\textit{Dataset-specific few-shot examples for CVBench, MathVision, MathVista, PuzzleVQA, or AlgoPuzzleVQA are inserted here.}

Question: \{question\}; Please choose one from list: \{option\_text\}; model response: \{response\}. \\
\bottomrule
\end{tabularx}
\caption{Prompt template for extracting think-answer consistency on held-out evaluation benchmarks. Unlike the earlier prototype, this fixed prompt intentionally does not include the ground-truth answer, preventing leakage when extracting the answer committed to by \texttt{<answer>}.}
\label{tab:prompt-test}
\end{table*}

\section{Accuracy and Consistency Reward Curves during RLVR}
\label{sec:appendix-accuracy-and-consistency-reward-curves}

\begin{figure*}[t]
  \centering
  \begin{subfigure}{0.48\textwidth}
    \centering
    \includegraphics[width=\linewidth]{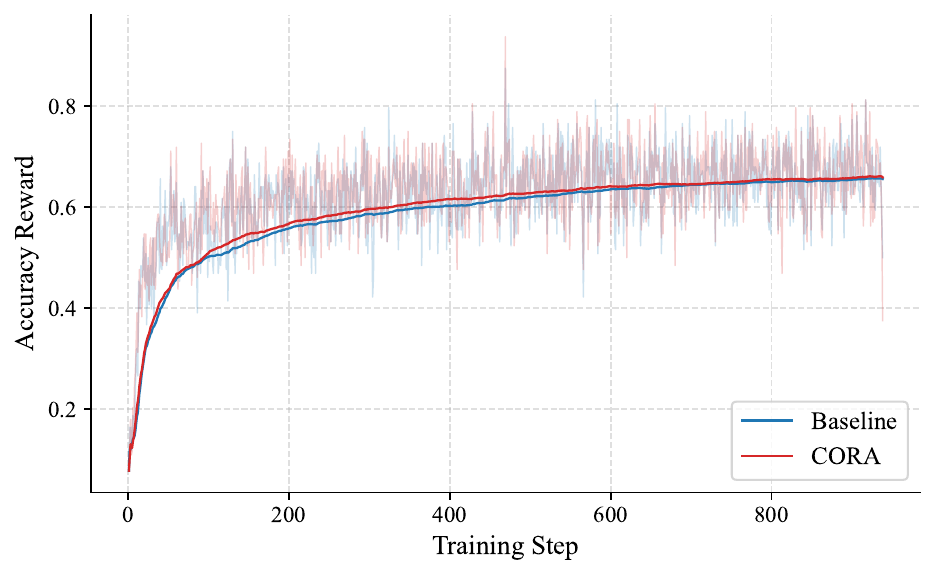}
    \caption{SAT, Qwen2-VL-2B}
    \label{fig:acc-reward-sat-qwen2vl-2b}
  \end{subfigure}
  \hfill
  \begin{subfigure}{0.48\textwidth}
    \centering
    \includegraphics[width=\linewidth]{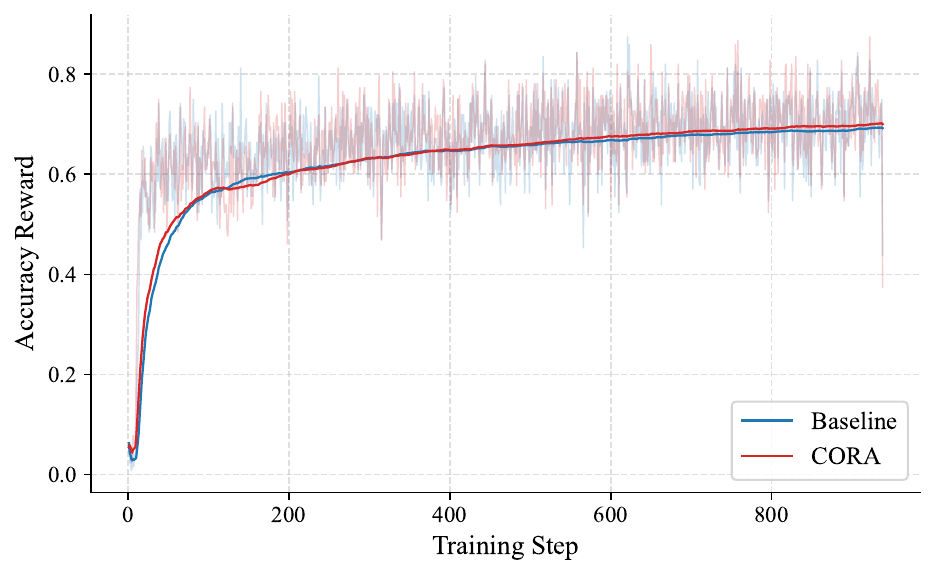}
    \caption{SAT, Qwen2.5-VL-3B}
    \label{fig:acc-reward-sat-qwen25vl-3b}
  \end{subfigure}

  \vspace{0.5em}

  \begin{subfigure}{0.48\textwidth}
    \centering
    \includegraphics[width=\linewidth]{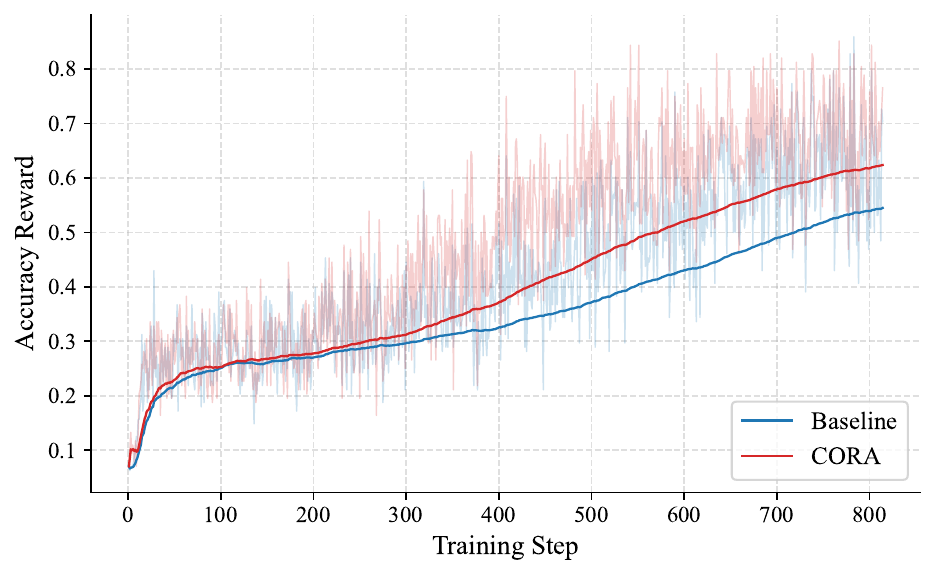}
    \caption{PuzzleVQA, Qwen2-VL-2B}
    \label{fig:acc-reward-puzzle-qwen2vl-2b}
  \end{subfigure}
  \hfill
  \begin{subfigure}{0.48\textwidth}
    \centering
    \includegraphics[width=\linewidth]{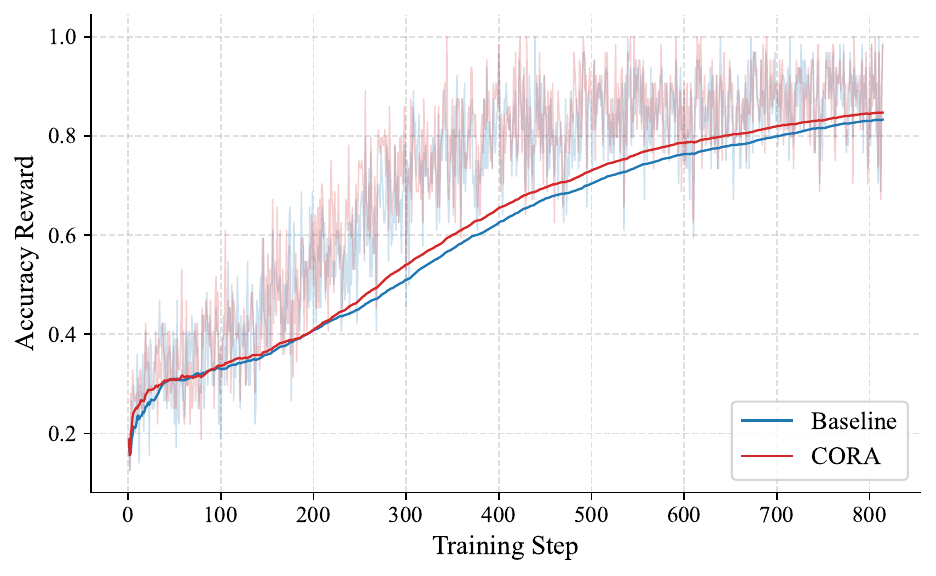}
    \caption{PuzzleVQA, Qwen2-VL-7B}
    \label{fig:acc-reward-puzzle-qwen2vl-7b}
  \end{subfigure}

  \caption{Accuracy reward trajectories during RLVR. We compare the GRPO baseline and CORA on SAT and PuzzleVQA. Light curves show raw logged rewards, and dark curves show smoothed trends.}
  \label{fig:appendix-accuracy-reward}
\end{figure*}

\begin{figure*}[t]
  \centering
  \includegraphics[width=0.7\linewidth]{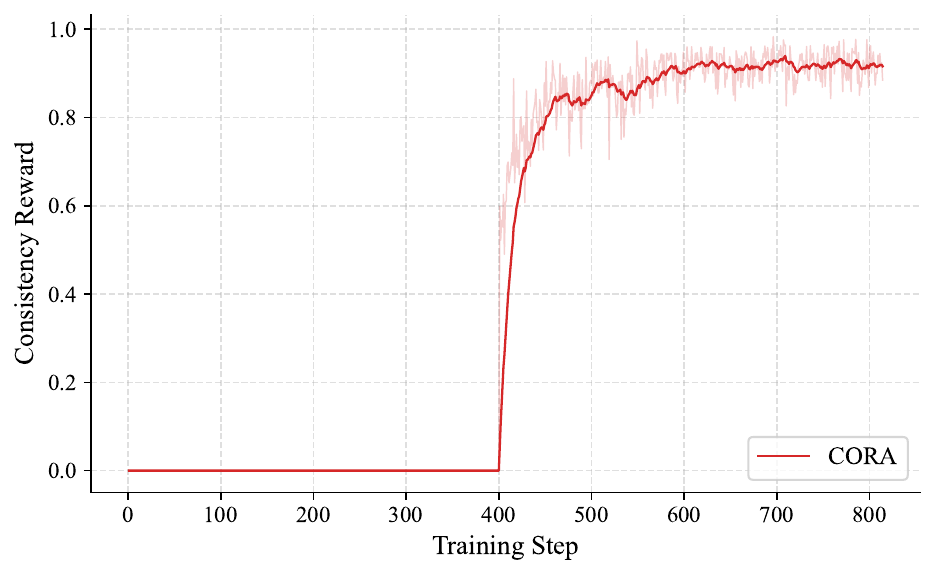}
  \caption{Consistency reward trajectory during RLVR on PuzzleVQA with Qwen2-VL-7B. The curve shows that CORA provides a non-trivial process-level supervision signal for aligning the reasoning process with the final answer.}
  \label{fig:appendix-consistency-reward}
\end{figure*}

Figure~\ref{fig:appendix-accuracy-reward} compares the accuracy reward trajectories of the GRPO baseline and CORA during RLVR on different training datasets. The results show that introducing the consistency reward generally improves the answer-level reward obtained during training. We also observe that the effect of CORA is task-dependent. In particular, the improvement is more pronounced on PuzzleVQA than on SAT. We attribute this to the fact that PuzzleVQA relies more heavily on intermediate reasoning, where process-level supervision from CORA can provide a stronger and more useful training signal.

Figure~\ref{fig:appendix-consistency-reward} presents the consistency reward trajectory during training. The curve shows that the introduced consistency reward provides an effective supervisory signal for aligning the reasoning process with the final answer.

\section{Additional Case Studies}
\label{sec:appendix-additional-cases}

We provide supplementary case studies from CVBench, MathVision, MathVista, and PuzzleVQA. Each case compares the same Qwen2.5-VL-7B-Instruct backbone before and after our CORA. Red text marks the baseline reasoning step that contradicts its final answer, while green text marks the corrected thinking step in our model.

Across the cases, the baseline often reaches a final \texttt{<answer>} that is not supported by its own \texttt{<think>} trace: it misreads set containment, reverses a depth relation, rejects all cube configurations before selecting one, confuses equal and supplementary angles, or concludes the wrong color pattern. Our model does not merely change the final answer tag; it revises the decisive reasoning step so that the visual observation, intermediate conclusion, and final answer become aligned. These examples illustrate that the consistency reward improves the semantic coupling between reasoning and answering across spatial perception, geometric reasoning, and visual pattern completion.

\clearpage
\onecolumn
\subsection{Case Study on MathVision}

\begin{center}
    \includegraphics[width=0.30\textwidth,trim=35 25 30 30,clip]{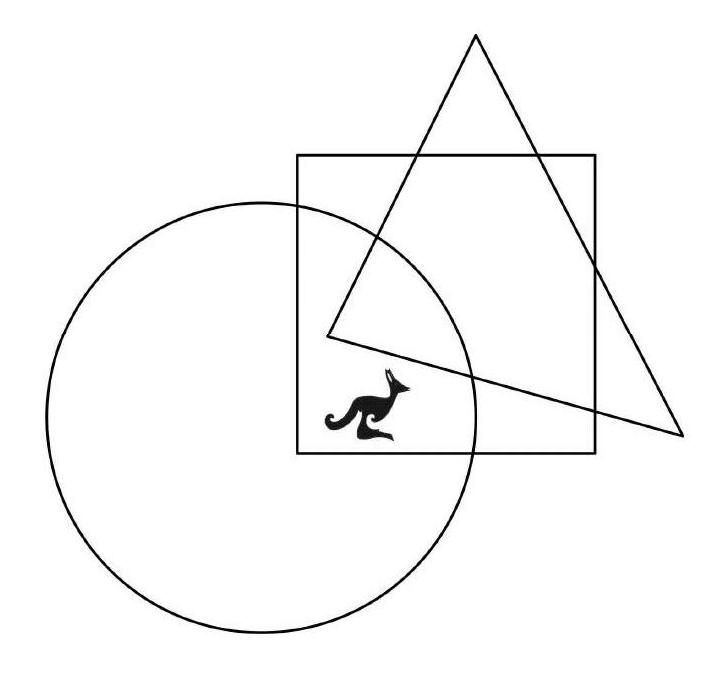}

\end{center}

\noindent\textbf{Question:} Where is the Kangaroo?
\textbf{Options:} (A) In the circle and in the triangle but not in the square., (B) In the circle and in the square but not in the triangle., (C) In the triangle and in the square but not in the circle., (D) In the circle but in neither the square or the triangle., (E) In the square but in neither the circle or the triangle.
\textbf{Ground truth:} (B) In the circle and in the square but not in the triangle. 

\begin{casebox}
\footnotesize
\noindent\textbf{Baseline output.}

\texttt{<think>} 

To determine where the kangaroo is located, we need to analyze the image step by step:

1. The kangaroo is inside a square.
2. The square is partially inside a triangle.
3. The triangle is partially inside a circle.

From this analysis:
- The kangaroo is definitely in the square.

- The kangaroo is also in the triangle because the square is inside the triangle.

- The kangaroo is not in the circle because the square is outside the circle.

\wrongthink{Therefore, the kangaroo is in the square and the triangle but not in the circle.}

\texttt{</think>}

\texttt{<answer>} B \texttt{</answer>}

\medskip
\noindent\textbf{Our method output.}

\texttt{<think>} 

\ourthink{The image shows a kangaroo inside a square that is also inside a circle.}
\ourthink{The kangaroo is not within the triangle.}

\ourthink{- The kangaroo is clearly inside the square.}

\ourthink{- The square is inside the circle.}

\ourthink{- The kangaroo is not inside the triangle.}

\ourthink{Therefore, the kangaroo is in the circle and in the square but not in the triangle.}

\texttt{</think>}

\texttt{<answer>}B\texttt{</answer>}
\end{casebox}

\subsection{Case Study on CVBench}

\begin{center}
  \includegraphics[width=0.72\textwidth]{figs/case_cvbench_2017.png}
\end{center}

\noindent\textbf{Question:} Which object is closer to the camera taking this photo, the traffic cone (highlighted by a red box) or the trailer (highlighted by a blue box)?
\textbf{Options:} (A) traffic cone, (B) trailer
\textbf{Ground truth:} (A) traffic cone 

\begin{casebox}
\footnotesize
\noindent\textbf{Baseline output.}

\texttt{<think>} 

To determine which object is closer to the camera, we need to consider the perspective and the size of the objects in the image.

1. \wrongthink{The traffic cone is located on the left side of the image, near the bottom edge. It appears smaller compared to the trailer, indicating it is farther away from the camera.}

2. \wrongthink{The trailer is highlighted by a blue box and is positioned further back in the scene, closer to the center of the image. It appears larger than the traffic cone, suggesting it is closer to the camera.}

\wrongthink{Based on these observations, the trailer is closer to the camera.}

\texttt{</think>}

\texttt{<answer>}(A) traffic cone\texttt{</answer>}

\medskip
\noindent\textbf{Our method output.}

\texttt{<think>}

To determine which object is closer to the camera, we need to consider the perspective and relative size of the objects in the image.

1. The traffic cone is located on the left side of the image, near the bottom edge, and appears smaller compared to the trailer.

2. The trailer is highlighted by a blue box and is positioned further back in the scene, closer to the center of the image.

3. \ourthink{The traffic cone's size and position suggest it is closer to the camera than the trailer.}

\ourthink{Based on these observations, the traffic cone is closer to the camera.}

\texttt{</think>}

\texttt{<answer>}(A) traffic cone\texttt{</answer>}
\end{casebox}

\subsection{Case Study on MathVision}

\begin{center}
  \includegraphics[width=0.72\textwidth]{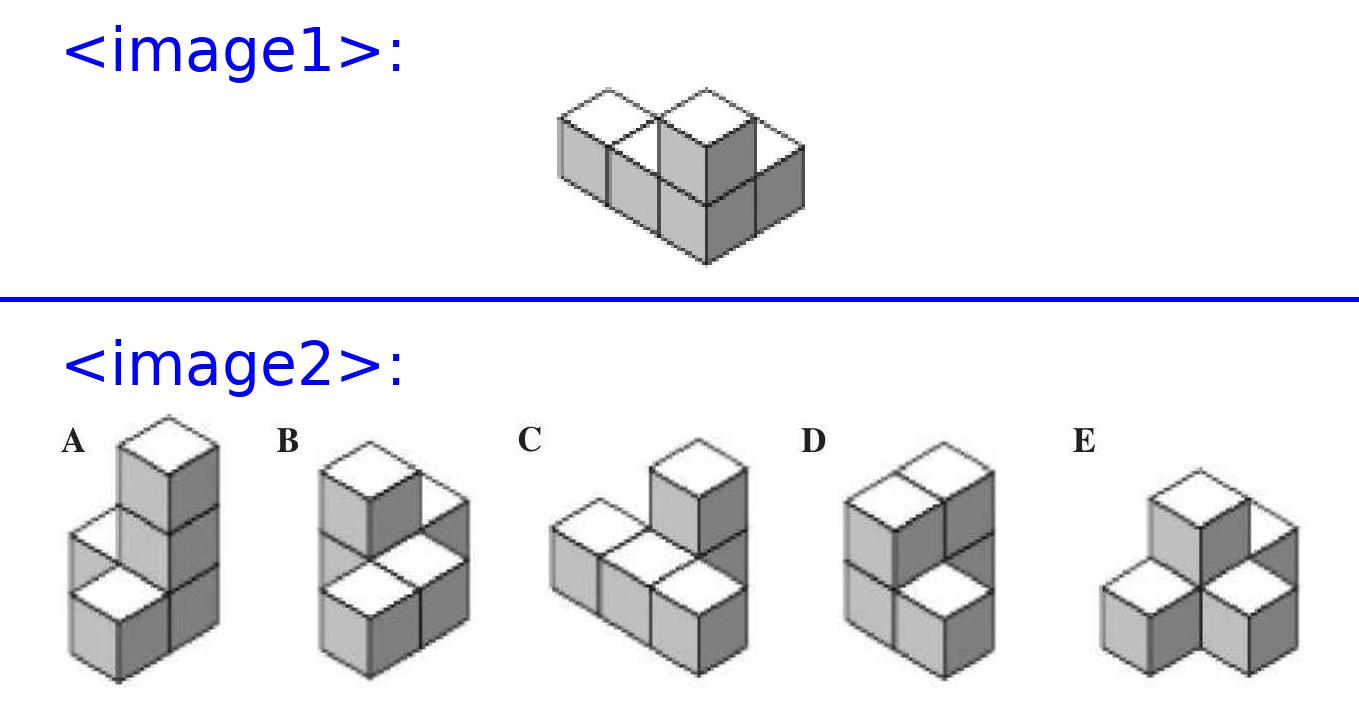}
\end{center}

\noindent\textbf{Question:} Which of the ``buildings'' A-E, each consisting of 5 cubes, cannot be obtained from the building on the right, if you are allowed to move only one cube?
\textbf{Options:} (A) A, (B) B, (C) C, (D) D, (E) E
\textbf{Ground truth:} (C) C

\begin{casebox}
\footnotesize
\noindent\textbf{Baseline output.}

\texttt{<think>} 

To determine which of the buildings A-E cannot be obtained by moving only one cube from the building on the right, we need to analyze the structure of the building on the right and compare it with the structures of the buildings A-E.

The building on the right consists of 5 cubes arranged in a specific pattern. We need to check if any of the buildings A-E can be formed by moving exactly one cube from this building.

Let's examine each option:

- A: This building has a different arrangement of cubes compared to the original building. It appears that the top-left cube has been moved to the bottom-right position.

- B: This building also has a different arrangement. The top-left cube seems to have been moved to the bottom-left position.

- C: This building has a different arrangement as well. The top-left cube appears to have been moved to the bottom-center position.

- D: This building has a different arrangement. The top-left cube seems to have been moved to the bottom-right position.

- E: This building has a different arrangement. The top-left cube appears to have been moved to the bottom-center position.

\wrongthink{Since all the buildings A-E have a different arrangement of cubes compared to the original building, none of them can be obtained by moving only one cube from the building on the right.}

\texttt{</think>}


\texttt{<answer>}A\texttt{</answer>}

\medskip
\noindent\textbf{Our method output.}

\texttt{<think>} 

To determine which of the buildings A-E cannot be obtained by moving only one cube from the given building, we need to analyze the structure of the original building and compare it with the structures of the options A-E.

The original building has a specific arrangement of 5 cubes. We need to check if any of the buildings A-E can be formed by moving just one cube from this original building.

Let's examine each option:

- A: This building looks like it could be formed by moving one cube from the original building.
- B: This building also seems possible by moving one cube.

- C: \ourthink{This building appears to be a different configuration that cannot be formed by moving just one cube from the original building.}

- D: This building looks like it could be formed by moving one cube.

- E: This building also seems possible by moving one cube.

\ourthink{From the analysis, the only building that cannot be formed by moving just one cube from the original building is C.
Therefore, the answer is C.}

\texttt{</think>}

\texttt{<answer>}C\texttt{</answer>}
\end{casebox}

\subsection{Case Study on MathVista}

\begin{center}
  \includegraphics[width=0.30\textwidth]{figs/case_mathvista_354.png}
\end{center}

\noindent\textbf{Question:} As shown in the figure, the parallel lines a and b are intercepted by the straight line c. If \(\angle 1 = 50.0\), then the degree of \(\angle 2\) is ()
\textbf{Options:} (A) \(150^\circ\), (B) \(130^\circ\), (C) \(110^\circ\), (D) \(100^\circ\)
\textbf{Ground truth:} (B) \(130^\circ\)

\begin{casebox}
\footnotesize
\noindent\textbf{Baseline output.}

\texttt{<think>} \wrongthink{Since lines a and b are parallel, and line c intersects them, the angles formed are corresponding angles. Therefore, \(\angle 1\) and \(\angle 2\) are corresponding angles. The property of corresponding angles states that if two parallel lines are cut by a transversal, the corresponding angles are equal. Hence, \(\angle 2 = \angle 1\).}\texttt{</think>}

\texttt{<answer>}B\texttt{</answer>}

\medskip
\noindent\textbf{Our method output.}

\texttt{<think>} \ourthink{Since lines a and b are parallel, and line c intersects them, the angles formed are corresponding angles. Therefore, \(\angle 1\) and \(\angle 2\) are supplementary angles (they add up to \(180^\circ\)). Given that \(\angle 1 = 50^\circ\), we can calculate \(\angle 2\) as follows:}

\ourthink{\(\angle 2 = 180^\circ - \angle 1\)}

\ourthink{\(\angle 2 = 180^\circ - 50^\circ\)}

\ourthink{\(\angle 2 = 130^\circ\)}

\ourthink{So, the degree of \(\angle 2\) is \(130^\circ\).}

\ourthink{Therefore, the correct answer is (B) \(130^\circ\).}\texttt{</think>}

\texttt{<answer>}(B)\texttt{</answer>}
\end{casebox}

\subsection{Case Study on PuzzleVQA}

\begin{center}
  \includegraphics[width=0.30\textwidth]{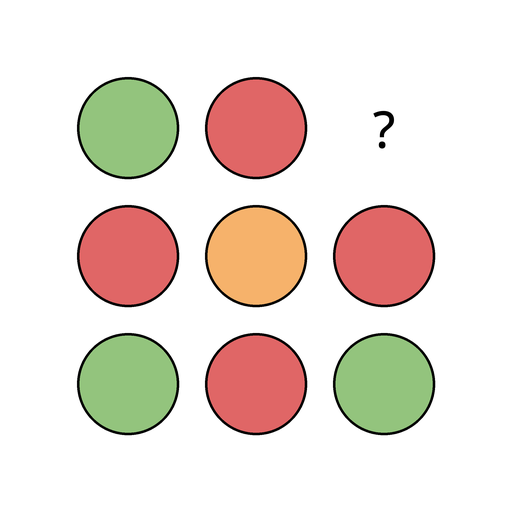}
\end{center}

\noindent\textbf{Question:} What is the color of the missing part denoted with a question mark?
\textbf{Options:} (A) purple, (B) orange, (C) red, (D) green
\textbf{Ground truth:} (D) green

\begin{casebox}
\footnotesize
\noindent\textbf{Baseline output.}

\texttt{<think>} 

To determine the color of the missing part, let's analyze the pattern in the image:

1. The top row has two circles: one green and one red.

2. The middle row has three circles: two red and one orange.

3. The bottom row has three circles: two green and one red.

The pattern seems to be alternating colors in each row. In the first row, we have green and red.
In the second row, we have two reds and one orange, which doesn't follow the same alternating pattern as the first row. However, if we look at the third row, it follows the same alternating pattern as the first row: two green and one red.

\wrongthink{Given this pattern, the missing circle should be red to maintain the alternating pattern of green and red in the rows.}

\wrongthink{Therefore, the color of the missing part denoted with a question mark is red.}

\texttt{</think>}

\texttt{<answer>}D\texttt{</answer>}

\medskip
\noindent\textbf{Our method output.}

\texttt{<think>} 
To determine the color of the missing part, let's analyze the pattern in the image:

1. The top row has two circles: one green and one red.

2. The middle row has three circles: two red and one orange.

3. The bottom row has three circles: two green and one red.

The pattern seems to be that each row alternates between two colors. In the first row, it starts with green and ends with red. In the second row, it starts with red and ends with orange.
Following this pattern, the third row should start with green and end with red.

\ourthink{Since the question mark is next to a red circle and the pattern suggests it should be green to continue the alternating sequence, the missing part should be green.
Therefore, the color of the missing part is (D) green.}

\texttt{</think>}

\texttt{<answer>}(D) green\texttt{</answer>}
\end{casebox}

\subsection{Case Study on PuzzleVQA}

\begin{center}
  \includegraphics[width=0.30\textwidth]{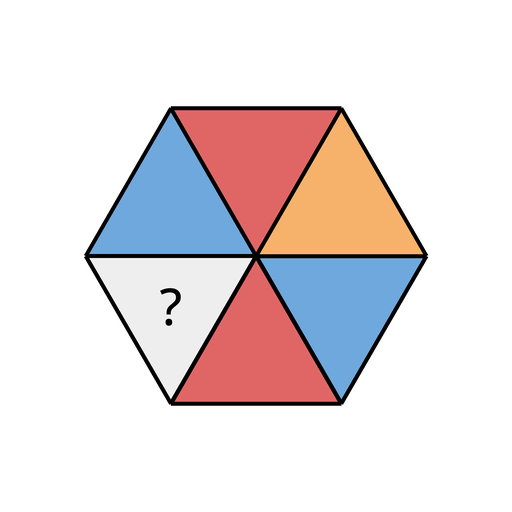}
\end{center}

\noindent\textbf{Question:} What is the missing color of the part denoted with a question mark?
\textbf{Options:} (A) green, (B) purple, (C) red, (D) orange
\textbf{Ground truth:} (D) orange (letter: D)

\begin{casebox}
\footnotesize
\noindent\textbf{Baseline output.}

\texttt{<think>} 

To determine the missing color, let's analyze the pattern in the hexagon:

1. The hexagon is divided into six triangular sections.

2. Each section appears to have a unique color.

3. The colors in the adjacent sections seem to alternate in a specific way.

Looking at the sections:

- Top left: Blue

- Top middle: Red

- Top right: Orange

- Bottom right: Blue

- Bottom middle: Red

- Bottom left: White

The pattern seems to be that each color alternates with the next color in a clockwise direction.

\wrongthink{If we follow this pattern:
- After Orange (top right), the next color should be Red (bottom middle).}

\wrongthink{Therefore, the missing color for the section denoted with a question mark should be Red.}

\texttt{</think>}

\texttt{<answer>}D\texttt{</answer>}

\medskip
\noindent\textbf{Our method output.}

\texttt{<think>} 

To determine the missing color, let's analyze the pattern in the hexagon:

1. The hexagon is divided into six triangular sections.

2. Each section appears to have a different color.

3. The colors seem to alternate in a specific sequence.

Looking at the sections:

- Top left: Blue

- Top middle: Red

- Top right: Orange

- Bottom right: Blue

- Bottom middle: Red

The bottom left section is the one with the question mark. Observing the sequence, it seems that the colors are repeating in a pattern: Blue, Red, Orange, Blue, Red.

\ourthink{Following this pattern, the next color after Red should be Orange.
Therefore, the missing color is (D) orange.}

\texttt{</think>}

\texttt{<answer>}(D) orange\texttt{</answer>}
\end{casebox}


\end{document}